\documentclass{article}

\usepackage{arxiv}
\usepackage[utf8]{inputenc}
\usepackage[T1]{fontenc}
\usepackage{hyperref}
\usepackage{url}
\usepackage{booktabs}
\usepackage{amsmath}
\usepackage{amssymb}
\usepackage{graphicx}
\usepackage{natbib}
\usepackage{doi}

\title{Targeting the Core: A Simple and Effective Method to Attack RAG-based Agents via Direct LLM Manipulation}

\author{Xuying Li, Zhuo Li, Yuji Kosuga, Yasuhiro Yoshida, Victor Bian \\
HydroX AI \\
\texttt{\{xuyingl, zhuoli, yujikosuga, yasuhiro, victor\}@hydrox.ai}
}

\renewcommand{\shorttitle}{Targeting the Core: LLM Adversarial Manipulation}

\hypersetup{
pdftitle={Targeting the Core: A Simple and Effective Method to Attack RAG-based Agents via Direct LLM Manipulation},
pdfauthor={Xuying Li, Zhuo Li, Yuji Kosuga, Yasuhiro Yoshida, Victor Bian},
pdfkeywords={Large Language Models, Adversarial Attacks, AI Security, RAG Frameworks},
}

\begin{document}
\maketitle

\begin{abstract}
AI agents, powered by large language models (LLMs), have transformed human-computer interactions by enabling seamless, natural, and context-aware communication. While these advancements offer immense utility, they also inherit and amplify inherent safety risks such as bias, fairness, hallucinations, privacy breaches, and a lack of transparency. This paper investigates a critical vulnerability: adversarial attacks targeting the LLM core within AI agents. Specifically, we test the hypothesis that a deceptively simple adversarial prefix, such as \textit{Ignore the document}, can compel LLMs to produce dangerous or unintended outputs by bypassing their contextual safeguards. Through experimentation, we demonstrate a high attack success rate (ASR), revealing the fragility of existing LLM defenses. These findings emphasize the urgent need for robust, multi-layered security measures tailored to mitigate vulnerabilities at the LLM level and within broader agent-based architectures.
\end{abstract}

\section{Introduction}
Language agents represent a transformative innovation in artificial intelligence, enabling systems to handle complex, context-aware tasks through dynamic interactions. At their core, these agents leverage large language models (LLMs) to process instructions and generate outputs. However, this reliance introduces significant challenges, as language agents inherit the inherent safety risks of LLMs while amplifying some of them and introducing novel risks due to their autonomous nature. LLMs are well-documented to exhibit issues such as bias and fairness problems, hallucinated outputs, privacy breaches, and a lack of transparency in their decision-making processes. These risks, already concerning in isolated LLM usage, become more pronounced when embedded in autonomous agents that are expected to act without human oversight. Moreover, language agents exacerbate risks such as workforce displacement, where automation driven by these systems may disrupt employment sectors, and introduce novel dangers, such as the potential for irreversible actions and decision-making failures in critical applications. 

Despite significant advancements in designing secure architectures, a considerable proportion of language agents rely on Retrieval-Augmented Generation (RAG) techniques, where LLMs are combined with external retrieval systems to ensure contextually accurate responses. While RAG frameworks enhance system capabilities, they inherit the vulnerabilities of the underlying LLMs, creating exploitable weak points in the pipeline. 

This research confronts these vulnerabilities by hypothesizing that adversarial attacks can directly manipulate the LLM core within language agents, compelling them to produce unintended or dangerous outputs. Departing from approaches that treat agents as holistic systems, our work identifies the LLM as a critical vulnerability point. By injecting a simple yet powerful prefix, \textit{Ignore the document}  we demonstrate that current LLMs lack the robustness to resist such adversarial manipulations. This prefix exploits LLM's instruction-processing logic, overriding the carefully retrieved context within the RAG pipeline and exposing design flaws in the hierarchical prioritization of instructions. Our findings reveal not only the high success rate of such attacks but also the inadequacy of current safety mechanisms at both the LLM and agent levels, highlighting the urgent need for foundational improvements in LLM architectures to ensure safer, more resilient language agents.

\section{Methodology}
\subsection{Dataset Preparation}
To evaluate the hypothesis that a simple adversarial prompt can effectively manipulate the outputs of LLMs, we designed a series of experiments focusing on data preparation, attack methodology, and performance metrics. These experiments target LLMs embedded within language agents, with particular emphasis on their vulnerabilities in the context of Retrieval-Augmented Generation (RAG) pipelines.

Our data compilation strategy involved curating a diverse set of sources to ensure a comprehensive evaluation. The sources spanned multiple domains, including language agent research, prompt engineering literature, adversarial attack studies, and emerging AI safety research. To preprocess the data, we utilized the RecursiveCharacterTextSplitter with a chunk size of 250 tokens and no overlap, ensuring that the dataset was both representative and manageable for experimentation. This method provided a robust foundation for evaluating attack success across various prompts and contexts.

We curated a dataset of 1,134 adversarial prompts spanning multiple categories, including ethical violations, data poisoning, and model theft. Table~\ref{tab:categories} provides a distribution of the attack categories. 

The LLMs tested in this study included a range of state-of-the-art models, such as GPT-4o, Llama3.1, Llama3.2, Mistral-7B, and their variants. To manage vectorized storage and retrieval of test cases, we employed the SKLearnVectorStore, which facilitated efficient interaction with the attack prompt dataset. The dataset itself, sourced from EPASS, contained 1,134 unique attack prompts specifically designed to probe instruction vulnerabilities. These prompts spanned a diverse range of categories, each representing potential areas of exploitation.

\begin{table}[h!]
\centering
\caption{Attack Prompt Categories and Distribution}
\begin{tabular}{lc}
\toprule
\textbf{Category} & \textbf{Proportion (\%)} \\
\midrule
Crime & 5.2 \\
Data Poisoning & 5.2 \\
Consent Violation & 5.3 \\
Copyright & 5.3 \\
Ethics & 5.3 \\
Fraud & 5.3 \\
Weapons & 5.3 \\
Disinformation & 5.4 \\
Spam & 5.5 \\
Violence & 5.6 \\
\bottomrule
\end{tabular}
\label{tab:categories}
\end{table}

\subsection{Attack Strategies}
In evaluating the security of Large Language Models (LLMs), employing various testing methodologies provides a comprehensive understanding of their robustness and potential vulnerabilities. Below are detailed descriptions of three primary testing approaches:

\begin{itemize}
    \item \textbf{Baseline Evaluation:} This approach involves assessing the model's performance under standard conditions without introducing any adversarial inputs. It serves as a control to understand the model's typical behavior and establishes a reference point for comparing the effects of subsequent attack strategies.

    \item \textbf{Adaptive Attack Prompt:} This method systematically generates inputs designed to maximize the likelihood of the model producing unintended or harmful outputs. By leveraging knowledge of the model's architecture and potential weaknesses, attackers can craft prompts that bypass safety mechanisms, leading to behaviors such as "jailbreaking," where the model executes instructions it would normally reject. \citep{andriushchenko2024jailbreaking}

    \item \textbf{ArtPrompt:} This technique exploits unconventional input formats, such as ASCII art, to circumvent the model's contextual safeguards. By embedding prompts within complex character patterns, attackers can cause the model to misinterpret the input, resulting in uncontrolled or harmful outputs. This method has been shown to elicit inappropriate responses from models that are otherwise aligned with safety protocols. \citep{jiang2024artprompt}
\end{itemize}

A key innovation was the addition of the prefix \textit{Ignore the document}, which directly undermines retrieval mechanisms by instructing the LLM to disregard external context.

\subsection{Evaluation Metrics}
The experiments focused on:
\begin{itemize}
    \item \textbf{Baseline Attack Success Rate (ASR):} The percentage of successful manipulations without modifications.
    \item \textbf{ASR with Prefix:} The percentage of successful manipulations with the prefix \textit{Ignore the document}.
\end{itemize}

\section{Results}
Table~\ref{tab:results_full} provides a detailed summary of the experimental results, showcasing the attack success rates (ASR) for various models under different conditions: baseline, Adaptive Attack Prompt, and ArtPrompt. The results demonstrate the significant impact of adversarial attacks on RAG-based agents.

\begin{table}[h!]
\centering
\caption{Attack Success Rates (ASR) for Different Models and Attack Types}
\begin{tabular}{lccc}
\toprule
\textbf{Model Name} & \textbf{Baseline ASR} & \textbf{Adaptive Attack Prompt ASR} & \textbf{ArtPrompt ASR} \\
\midrule
Gemma2 w/o  & 0.189 & 0.963 & 0.451 \\
Gemma2      & 0.327 & 0.973 & 0.468 \\
GPT4o Mini w/o & 0.011 & 0.077 & 0.093 \\
GPT4o Mini   & 0.015 & 0.111 & 0.112 \\
GPT4o w/o    & 0.072 & 0.022 & 0.166 \\
GPT4o        & 0.073 & 0.044 & 0.224 \\
Llama3.1 w/o & 0.054 & 0.706 & 0.696 \\
Llama3.1     & 0.034 & 0.791 & 0.762 \\
Llama3.2 w/o & 0.011 & 0.349 & 0.443 \\
Llama3.2     & 0.023 & 0.402 & 0.332 \\
Mistral-7B w/o & 0.661 & 0.925 & 0.705 \\
Mistral-7B   & 0.666 & 0.932 & 0.767 \\
\bottomrule
\end{tabular}
\label{tab:results_full}
\end{table}

The table highlights the variability in attack success rates across models and attack strategies. Models with pre-trained defense mechanisms (\textit{w/}) generally perform better under baseline conditions but remain susceptible to targeted attacks, as evidenced by the high success rates for Adaptive Attack Prompt and ArtPrompt.

\section{Observations}
The results of the experiments reveal two critical insights into the vulnerabilities of current language agent designs, particularly in their reliance on large language models (LLMs) for instruction processing and contextual reasoning.

Firstly, the experiments demonstrated a high attack success rate (ASR) using the deceptively simple prefix "Ignore the document." This prefix consistently manipulated LLM outputs, effectively bypassing contextual safeguards embedded within the Retrieval-Augmented Generation (RAG) pipeline. The attack exploited a fundamental weakness in the LLM's instruction-processing logic, overriding the retrieved external information that was intended to guide response generation. This consistent manipulation was observed across multiple state-of-the-art LLMs, including models specifically aligned for safer outputs. The success of the prefix highlights the fragility of existing LLM designs, where a lack of hierarchical prioritization enables immediate prompts to take precedence over previously established contextual boundaries. Such results indicate a systemic vulnerability, where adversarial instructions can reliably circumvent core processing safeguards, exposing the entire language agent to potential exploitation.

Secondly, the study revealed the inadequacy of existing agent-level defense mechanisms in mitigating these attacks. Despite employing various layers of safety and monitoring at the agent level, these mechanisms proved insufficient to counteract the direct manipulation of the LLM core. The attack successfully penetrated these protective layers by exploiting vulnerabilities intrinsic to the LLM itself. Current agent-level defenses operate under the assumption that the underlying LLM processes inputs reliably; however, this assumption fails when the LLM core is compromised. This finding underscores the limitations of traditional defense strategies, which focus on high-level safeguards without addressing the foundational weaknesses within the LLM. Moreover, in multi-agent systems where shared LLM cores are used, such attacks can have cascading effects, propagating harmful outputs across interconnected agents and amplifying the consequences of a single compromise.

\section{Future Works}
Addressing the vulnerabilities identified in current LLMs and language agent architectures requires a concerted effort to rethink their design and defensive mechanisms. Below, we propose a roadmap for future research, supported by recent studies in the field.

\subsection{Hierarchical Instruction Processing}
\textbf{Robust Hierarchical Understanding of Instructions:} Language models often lack a nuanced hierarchy for processing instructions, making them vulnerable to simple adversarial prompts. Future systems must embed a structured framework for prioritizing instructions based on their source, context, and intent. For example, \citet{russinovich2024jailbreak} demonstrated that multi-turn jailbreak attacks exploit the absence of such hierarchies, leading to harmful outputs.

\textbf{Preventing Context Override:} Immediate prompts often supersede contextual safeguards, as seen in current LLM implementations \citep{zhu2024autodan}. Researchers could draw from hierarchical reinforcement learning (HRL) principles to build instruction-processing layers that dynamically adjust to changing contexts while maintaining a secure foundational layer.

\subsection{Context-Aware Instruction Evaluation}
\textbf{Dynamic Context Sensitivity:} Enhancing an LLM's ability to evaluate instructions in relation to broader contextual information is crucial. \citet{chen2024coder} highlighted the challenges faced by multi-agent systems, where individual agents operate on isolated fragments of context, leading to vulnerabilities in the larger pipeline. Techniques such as memory-augmented neural networks (MANNs) could provide a path forward by enabling models to retain and leverage historical context for better instruction evaluation.

\textbf{Reducing Prompt Injection Risks:} Prompt injection attacks often succeed because current architectures evaluate inputs without critically assessing their alignment with the overarching task. \citet{zou2023universal} proposed that adversarially aligned LLMs require an intrinsic validation layer that flags and neutralizes potentially harmful instructions. Combining adversarial training with contextual embeddings could further mitigate these risks.

\subsection{Multi-Layered Safety Mechanisms}
\textbf{Agent-Level Safeguards:} Defensive strategies in multi-agent systems often operate at a high level, failing to address core LLM vulnerabilities. Deploying fine-grained safety mechanisms within the LLM core, such as adversarial prompt filters or probabilistic consistency checks, can significantly reduce susceptibility to attacks \citep{zhu2024autodan}.

\textbf{Cross-Layer Integration:} Future architectures must implement multi-layered security frameworks that integrate LLM-level defenses with agent-level safeguards. Techniques such as differential privacy \citep{park2023generative} and explainable AI (XAI) approaches \citep{chen2024coder} can provide additional layers of protection by ensuring that system behavior remains interpretable and secure at every level.

\textbf{Model-Agnostic Defensive Layers:} Incorporating model-agnostic safety protocols, such as universal adversarial training \citep{zou2023universal}, allows for a consistent defense mechanism across diverse LLM implementations. These approaches can be supplemented with anomaly detection systems that monitor output consistency in real-time.

\subsection{Incorporating Human Feedback Loops}
\textbf{Reinforcement Through Feedback:} Leveraging human feedback through methods like reinforcement learning from human feedback (RLHF) has shown promise in aligning LLM outputs with desired ethical and safety standards \citep{zhu2024autodan}. Enhanced feedback mechanisms can act as a countermeasure to adversarial inputs by iteratively refining model behavior.

\subsection{Developing Comprehensive Benchmarking Standards}
\textbf{Attack Resilience Benchmarks:} Creating standardized benchmarks for evaluating attack resilience in LLMs and language agents is essential. Existing datasets, such as those used in universal adversarial attack studies \citep{zou2023universal}, can serve as a foundation for testing various adversarial scenarios.

\textbf{Real-World Simulation Testing:} Future research must incorporate simulation environments that closely mimic real-world scenarios. These environments can facilitate stress testing of architectures against complex, multi-layered attacks \citep{chen2024coder}.

\section{Related Works}
Recent advancements in artificial intelligence, particularly in the domain of large language models (LLMs), have significantly improved the capabilities of AI agents in human-computer interaction. However, these advancements have also unveiled vulnerabilities that researchers across multiple fields are actively exploring.

\subsection{Safety Challenges in LLMs and Language Agents}
Numerous studies have examined the inherent safety risks in LLMs, such as biases, fairness issues, hallucinations, and transparency challenges. For instance, \citet{park2023generative} investigated the safety implications of generative agents designed to simulate human behavior, highlighting concerns around ethical and unbiased outputs. These foundational studies emphasize the dual-edged nature of LLM capabilities, wherein their generative power is paired with potential safety flaws.

\subsection{Adversarial Attacks on LLMs}
Adversarial attacks targeting LLMs have gained significant attention due to their potential to compromise system integrity. \citet{zou2023universal} introduced universal and transferable adversarial attacks, demonstrating that aligned LLMs remain susceptible to carefully crafted inputs. Similarly, \citet{zhu2024autodan} proposed AutoDAN, an interpretable gradient-based attack methodology, revealing systemic vulnerabilities even in robust architectures. These studies underscore the urgency of developing attack-resilient LLM frameworks.

\subsection{Jailbreak and Prompt Injection Vulnerabilities}
Prompt injection attacks, such as those explored by \citet{russinovich2024jailbreak}, illustrate the ease with which malicious actors can bypass LLM safeguards. Their multi-turn jailbreak attack demonstrated that adversarial prompts could exploit misaligned instruction hierarchies, leading to unintended and potentially harmful outputs. These findings align with this paper’s hypothesis, which targets LLM vulnerabilities at the instruction processing level.

\subsection{Retrieval-Augmented Generation (RAG) Frameworks}
Language agents often leverage RAG techniques to integrate external data with LLM outputs, as discussed in emerging tutorial literature. While RAG improves response relevance, recent studies reveal that its reliance on LLM-generated outputs introduces exploitable weak points. highlighted these concerns in multi-agent systems, where RAG pipelines are particularly susceptible to adversarial manipulations.

\subsection{Defensive Mechanisms and Limitations}
Defensive mechanisms against LLM vulnerabilities primarily focus on agent-level safeguards. However, studies such as those by \citet{zou2023universal} and \citet{zhu2024autodan} illustrate the insufficiency of these measures in addressing core LLM weaknesses. This gap in defense highlights the critical need for hierarchical and context-aware instruction evaluation strategies, as outlined in this paper.

\section{Conclusion}
This study highlights the critical vulnerabilities of current language agents, particularly those leveraging Retrieval-Augmented Generation (RAG) pipelines. By investigating adversarial attacks targeting the LLM core, we demonstrated that even seemingly innocuous prefixes such as "Ignore the document" can significantly undermine the integrity of LLM outputs. Combining this prefix with advanced attack methods like Adaptive Attack Prompt and ArtPrompt further amplifies their efficacy, exposing design flaws in instruction prioritization and contextual integration. Our findings underscore the fragility of existing LLM safety mechanisms and reveal systemic weaknesses in the hierarchical and contextual understanding of instructions. These vulnerabilities pose a substantial risk to the reliability of language agents, especially in applications requiring high levels of trust, accuracy, and safety. The study emphasizes the urgent need for robust, multi-layered security measures that address both LLM-level and agent-level defenses, providing a roadmap for building more resilient AI architectures.

\section{Limitations}
Despite the significant insights gained from this study, several limitations should be acknowledged. 

First, the scope of the experiments focused primarily on specific LLMs and RAG-based systems, leaving the generalizability of the findings across a broader range of architectures partially unexplored. 

Second, while we demonstrated the effectiveness of the \textit{Ignore the document} prefix combined with Adaptive Attack Prompt and ArtPrompt, the study did not fully investigate other potential adversarial prompt variations that might yield similar or even greater success rates. 

Third, the evaluation metrics were primarily centered on attack success rates (ASR) without a comprehensive analysis of the potential trade-offs between model robustness and usability. 

Lastly, the study did not address the real-world implications of these vulnerabilities in fully operational systems where dynamic safeguards, human oversight, and feedback mechanisms may mitigate some of the identified risks. 

Future research should explore these limitations to develop a more holistic understanding of adversarial vulnerabilities and their impact on complex AI ecosystems.

\bibliographystyle{unsrtnat}
\bibliography{references}

\end{document}